\documentclass[lettersize,journal]{IEEEtran}
\usepackage{amsmath,amsfonts}
\usepackage{algorithmic}
\usepackage{algorithm}
\usepackage{array}
\usepackage[caption=false,font=normalsize,labelfont=sf,textfont=sf]{subfig}
\usepackage{textcomp}
\usepackage{stfloats}
\usepackage{url}
\usepackage{verbatim}
\usepackage{graphicx}
\usepackage{cite}
\usepackage[numbers]{natbib}

\usepackage{hyperref}
\begin{document}

\title{Continual Learning in Medical Imaging: A Survey and Practical Analysis}


\author{
  Mohammad Areeb Qazi$^{1}$, 
  Anees Ur Rehman Hashmi$^{1}$, 
  Santosh Sanjeev$^{1}$, 
  Ibrahim Almakky$^{1}$, 
  Numan Saeed$^{1}$, 
  Camila Gonzalez$^{2}$, 
  Mohammad Yaqub$^{1}$ \\
  $^{1}$Mohamed bin Zayed University of Artificial Intelligence, Abu Dhabi, UAE \\
  $^{2}$Stanford University, USA
}

\maketitle

\begin{abstract}
Deep Learning has shown great success in reshaping medical imaging, yet it faces numerous challenges hindering widespread application. Issues like catastrophic forgetting and distribution shifts in the continuously evolving data stream increase the gap between research and applications. Continual Learning offers promise in addressing these hurdles by enabling the sequential acquisition of new knowledge without forgetting previous learnings in neural networks. In this survey, we comprehensively review the recent literature on continual learning in the medical domain, highlight recent trends, and point out the practical issues. Specifically, we survey the continual learning studies on classification, segmentation, detection, and other tasks in the medical domain. Furthermore, we develop a taxonomy for the reviewed studies, identify the challenges, and provide insights to overcome them. We also critically discuss the current state of continual learning in medical imaging, including identifying open problems and outlining promising future directions. We hope this survey will provide researchers with a useful overview of the developments in the field and will further increase interest in the community. To keep up with the fast-paced advancements in this field, we plan to routinely update the repository with the latest relevant papers at \href{https://github.com/BioMedIA-MBZUAI/awesome-cl-in-medical}{https://github.com/BioMedIA-MBZUAI/awesome-cl-in-medical}
\end{abstract}

\begin{IEEEkeywords}
Continual Learning, Medical, Survey, Types of Conntinal Learning
\end{IEEEkeywords}

\section{Introduction}
\label{intro}

\begin{figure*}[!t]
\centering
\includegraphics[width=\linewidth]{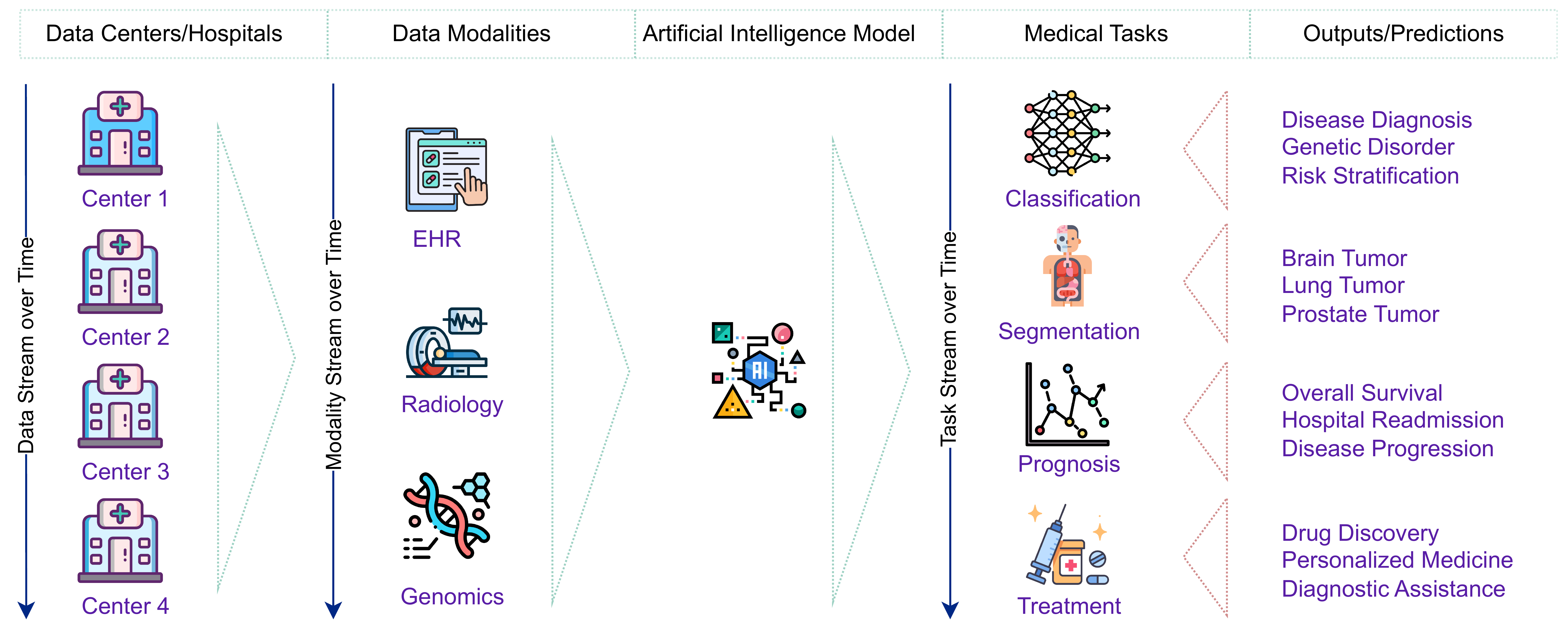}
\caption{The different changes through time that can occur in a healthcare scenario. The changes are depicted horizontally, with the time flow depicted vertically. The aim in CL is for the AI model depicted in the center of the figure to adapt to these data changes through time. Updates can include (i) a new center, (ii) a new modality, and/or (iii) a new task}
\label{fig:cl_full}
\end{figure*}

Deep Learning (DL) algorithms are rapidly gaining relevance in medical imaging, enabling segmentation \cite{chen2020deep,hofmanninger2020automatic}, classification \cite{kumar2023medical}, and detection \cite{halder2020lung} of anatomical structures and anomalies \cite{schlegl2019f,qazi2023multi} relevant for diagnosis, prediction, or prognosis. In some cases, their capabilities surpass those of human experts \citep{walsh2018deep,de2018clinically}, making them a central tool in the advancement of the use of medical data to support clinical decisions \citep{ahsan2022machine, adlung2021machine}. Although there are several DL works in medicine, the task is often oversimplified by assuming a static, centralized data source and neglecting the dynamic nature of the evolving data.

In general, a model is trained and tested on different splits of the same dataset, and the models are expected to generalize well to other datasets of the same nature. However, this does not reflect the reality of real-world deployment because of frequent changes in data distribution across various institutions, diagnostic devices, and population demographics \citep{cohen2021problems}. For Instance, in 2020, Google developed a DL system trained on a vast heterogeneous dataset that achieved a high performance in an offline setting for detecting diabetic eye diseases \citep{beede2020human}. However, when this system was deployed, its performance drops drastically, highlighting the severity of the problem and emphasizing the need for robust systems. These issues can be partially addressed through domain adaptation and generalization.

In real-world scenarios, a model trained at a specific point in time struggles to maintain optimal performance owing to the continuous evolution of the data distributions. Retraining the model on the entire dataset is impractical because substantial computational resources are required, and privacy concerns often restrict access to previous data, exacerbating this issue. Additionally, sequential fine-tuning of models can lead to overfitting of the latest task, resulting in a decline in the performance on previous data. 

Continual learning (CL) has emerged as a promising approach for training deep neural networks in order to overcome these challenges. CL offers a solution that enables neural networks to learn continuously, resulting in a more resilient model capable of accommodating new tasks without forgetting previous tasks. Although the model remains static when deployed in a static environment, it continuously updates itself in a CL setting, thereby yielding a more robust and versatile model. Addressing the challenges mentioned above requires consideration of the rigidity and plasticity of the neural networks. Rigidity, the ability of a neural network to preserve previous knowledge, and plasticity, the ability to learn new tasks are crucial factors. A model with high plasticity can learn new tasks over time, achieving good performance on future tasks but deteriorating performance on older tasks, a phenomenon known as \textit{"catastrophic forgetting"}. An ideal neural network should balance rigidity and plasticity to demonstrate its ability to learn new tasks without forgetting prior knowledge. CL has emerged as a promising approach for balancing these properties, enabling neural networks to learn continuously and adapt to evolving data distribution.

\subsection{Motivation}

\begin{figure}[b]
\centering
\includegraphics[width=\linewidth]{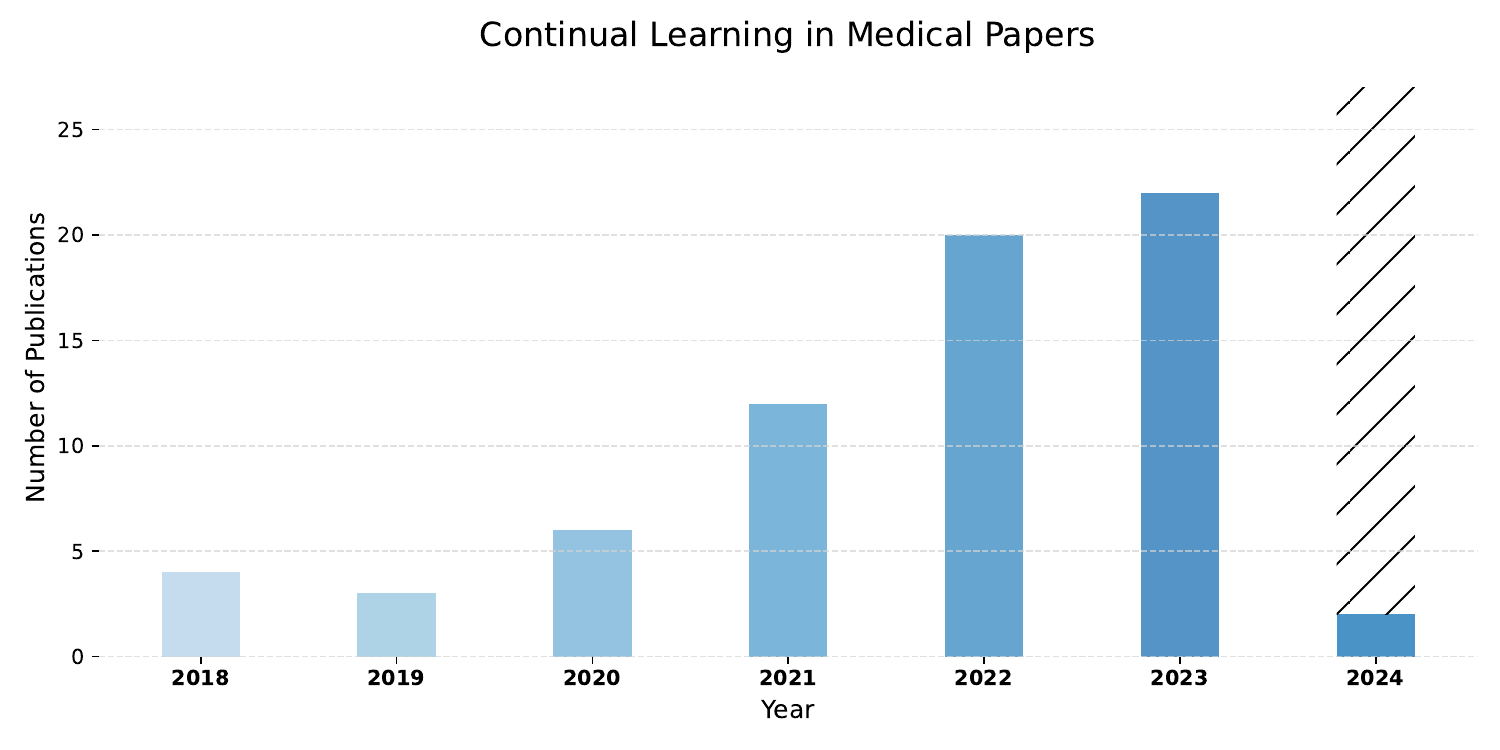}
\caption{Trend of studies in Continual Learning for Medical Imaging between 2018 and 2023 with the current and projected studies from 2024.}
\label{fig:trend_timeline}
\end{figure}

Significant efforts have been made to develop CL techniques for the natural and medical imaging domains \citep{douillard2022dytox,wang2022learning,wang2022dualprompt,smith2023coda,qazi2024dynammo}. However, numerous unique challenges arise when dealing with medical data. 
As illustrated in Figure  \ref{fig:cl_full}, medical imaging often encounters the challenge of changes in data distribution over time due to both unexpected factors and significant changes in tasks or modalities. Unexpected factors include (a) the incorporation of data from new medical centers, which might vary in acquisition equipment, image reconstruction algorithms, population demographics, and disease expression, and (b) changes within the same site, such as equipment degradation or replacement. Significant changes, on the other hand, include (c) the integration of new modality streams that could enhance downstream tasks and (d) the addition of new tasks, which may require different approaches. Several studies have been conducted to address these challenges. In this work, we aim to conduct a survey of the existing literature and research that may highlight the solutions, the practicality of the solutions, and the trend of contributions in CL in medicine. Several surveys have been conducted on CL using natural images \citep{van2022three,wang2023comprehensive,zhou2023deep}. However, little attention has been paid to the field related to CL and its applications in medicine, a broad and important research area. Moreover, as the trend increases, as shown in Figure \ref{fig:trend_timeline}, a detailed review of this field becomes necessary. Consequently, a survey that captures all aspects in the field is essential. 

\subsection{Study Selection}

Our selection process involved an extensive search targeting studies centered on the application of CL in medical settings. We initially identified 2100 studies by searching for continual learning and medicine from various databases and digital libraries including Scopus, Springer and ScienceDirect. Subsequently, we refined our focus to 177 relevant computer science and engineering papers from 2018 onwards. These papers were then assessed based on the inclusion and exclusion criteria, where we evaluated their alignment with our research objectives. The inclusion criteria were as follows: (a) addressed both CL and Medical Imaging; (b) proposed solutions for clinical challenges in the medical field; and (c) offered insights or remedies concerning issues related to catastrophic forgetting. Conversely, the exclusion criteria eliminated papers that (a) solely focused on CL without relevance to medical imaging, (b) primarily concentrated on FDA-related topics, and (c) addressed only catastrophic forgetting without relevance to medical imaging or clinical problem solving. Following the study selection process, we curated the final selection of 67 papers. Our final list of studies includes papers that not only explored essential aspects of continual learning in medical imaging but also offered significant insights and solutions for clinical challenges while addressing issues related to catastrophic forgetting.

\subsection{Paper  Organisation}
The rest of this paper is organized as follows. We start by mathematically formalizing CL in Section \ref{backgroud}. Then, we discuss and formulate the three types of CL: Task Incremental Learning (Section \ref{TIL}), Domain Incremental Learning (Section \ref{DIL}), and Class Incremental Learning (Section \ref{CIL}). Within this section, we first discuss the theory of these three types of CL before mathematically formulating them with an example. In Section \ref{taxonomy}, we review the recent advances in CL methods in the medical field and divide the studies on the approach they took to alleviate forgetting. Then, in Section \ref{discussion}, we discussed the overview of the studies along with the current research trend (section \ref{trend}) and the practicality of these studies (section \ref{practicality}). Section \ref{recommendation} provides insight into future directions in the field and section \ref{conclusion} concludes the paper.

\section{Background}
\label{backgroud}

\begin{figure*}[t]
\centering
\includegraphics[width=0.85\linewidth]{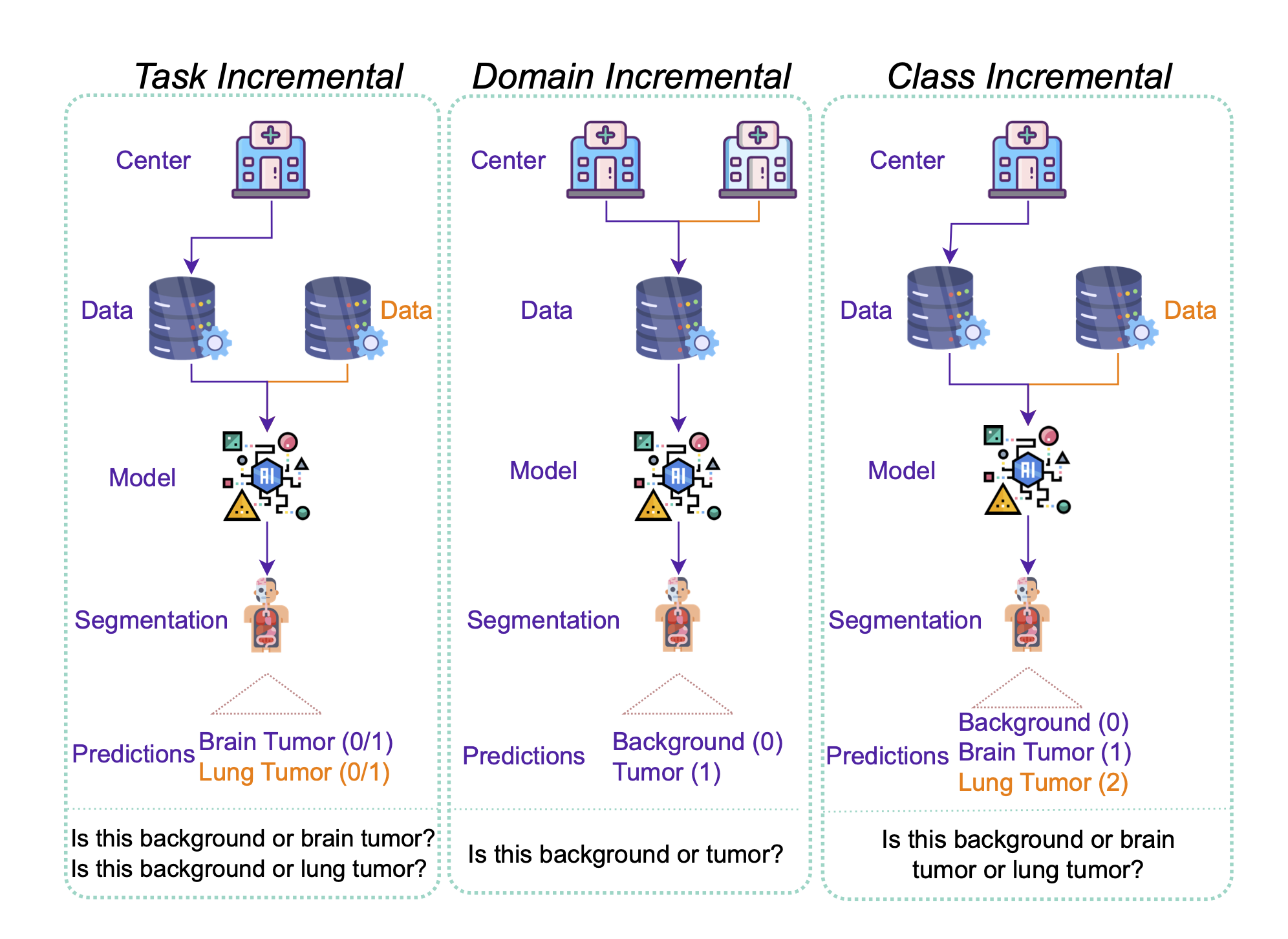}
\caption{The Three Types of Continual Learning. While Class-Incremental Learning and Task-Incremental Learning share training methodologies, Task-Incremental Learning requires task ID at inference. Meanwhile, Domain-Incremental Learning maintains consistent labels across tasks with varying data distributions.}
\label{fig:cl_type}
\end{figure*}

In this section, we present a comprehensive conceptual framework that includes the mathematical formulation of three primary types of CL: task, domain, and class incremental learning. The framework highlights the potential of these three types and how they can be applied to various advanced medical tasks. The purpose of this framework is to underline the essential concepts of CL, and how diagnostic and prognostic models with CL capabilities are robust and adaptable. The framework showcases the applications of CL in the medical field through Figure \ref{fig:cl_type}, thereby highlighting CL as a foundational feature in models that need to incrementally learn from new data sources while retaining previously learned knowledge. Then we summarize common evaluation metrics used for continual learning methods.

\subsection{Types of Incremental Learning}

To formalize the CL framework, we consider a model \textit{M} represented by a parametric function \textit{f}, which aims to learn the mapping between the input space \textit{X} and the output space \textit{Y}. Here, \textit{X} represents the medical data, including sensor readings, images, and electronic health records (EHR), while \textit{Y} includes diagnostic classifications, tumor or organ segmentation, and other medical-related predictions. Function \textit{f} is parameterized by weights $\theta$. In a CL setting, \textit{f} must adapt these parameters as new data becomes available, such as tasks, domains, and/or classes.

One of the biggest challenges in CL is the phenomenon of catastrophic forgetting, in which learning new information leads to forgetting previously known knowledge. The main objective is to optimize the parameters $\theta$ of the function \textit{f} over a data distribution \textbf{D}, which changes over time, reflecting new tasks (\textit{T}), domains (\textit{D}), or classes (\textit{C}).

\begin{equation}
     \arg \min_{\theta_{t+1}} \left[\mathcal{L}\left(f\left(X_{(t)}, \theta_{t+1}\right), Y_{(t)}\right) + \lambda \Omega(\theta_{t+1}, \theta_t)  \right] 
\end{equation}

\noindent where \textit{t} represents the time, \textit{X$_t$} and \textit{Y$_t$} are the input and output at time \textit{t}, the loss function $\mathcal{L}$ calculates the difference between the model prediction and the true outputs, $\Omega$ penalizes the change in parameters important for the previously learned knowledge, and $\lambda$ is a hyperparameter that balances the retention of old knowledge and new learning. 

In the remainder of this section, we use this mathematical framework to formalize the three types of CL techniques. Each presents its own challenges and requires a tailored approach to alleviate forgetting and seamlessly learning new knowledge. This shared vocabulary based on mathematical formulation will help communicate solutions for CL problems. 

\subsubsection{Task-Incremental Learning}
\label{TIL}

\textit{Theory:} In the scope of CL, task-incremental learning refers to the model's ability to learn new tasks sequentially, using the previously learned knowledge. This method emphasizes on clear separation of tasks with distinct objectives. It is always explicitly defined for the model what task to perform by presenting a task identifier, thereby avoiding any interference between tasks. This approach allows the model to incrementally learn new tasks without forgetting the old ones. 

\textit{Formulation:} Formally, given a non-stationary \textit{n} tasks ${T_1, T_2, T_3, ..., T_n}$, that require to learn the mapping from $X_i$ to $Y_i$ for each task, the task-incremental protocol relies on the task identifier $t$ to distinguish between the tasks, and thereby, performing the tasks individually. In this design, the model can be developed to include, for example, an individual output layer for each task or include task-distinctive components in the network. Mathematically, this can be presented as optimizing the following objective function for each task:

\begin{equation}
    \arg \min_{\theta_{\tau+1}} \left[ \mathcal{L}\left(f\left(X_{T_i}, \theta_{\tau+1}\right), Y_{T_i}\right) + \lambda \Omega\left(\theta_{\tau+1}, \theta_{\tau}\right) \right]
\end{equation}

\noindent where $\theta_{\tau}$ are the parameters of the model-specific task $\tau$, $\mathcal{L}$ is the loss function and $\Omega$ is the regularization term that conserves the knowledge from previously learned tasks.

\textit{Example:} Figure \ref{fig:cl_type} (left) provides an illustration of task-incremental learning in medical field. Initially, the model is trained to segment brain tumors. As the new data arrives, the model learns to identify lung tumors, using a new set of parameters while retaining the knowledge from previously learned brain tumor segmentation tasks. This sequential learning process is represented in the same model by two separate pathways, each activated by its corresponding task identifier.

\subsubsection{Domain-Incremental Learning}
\label{DIL}

\textit{Theory:} The domain-incremental learning is characterized by the model's capability to generalize across different domains or contexts while incorporating new data without explicit identifiers. This type of learning is critical when the underlying task is the same, but the data arrives incrementally from different sources. The different sources can have unique characteristics, such as imaging data from different scanners or demographics.

\textit{Formulation:} Formally, for a sequence of domains $D_1$ through $D_n$, the aim is to learn the mapping from $X_i$ to $Y_i$ for each domain. As the domain incremental approach performs the task without any information about the sample domain (context) and the output is similar for each domain, hence no identification of the context is required.  The domain-incremental learning approach tunes the parameters $\theta$ to optimize the following loss across domains:

\begin{equation}
    \arg \min_{\theta_{\varphi+1}} \left[ \mathcal{L}\left(f\left(X_{D_i}, \theta_{\varphi+1}\right), Y_{D_i}\right) + \lambda \Omega\left(\theta_{\varphi+1}, \theta_{\varphi}\right) \right]
\end{equation}

\noindent where $\theta_{\varphi}$ are the parameters of the model-specific domain $\varphi$, $\mathcal{L}$ is the loss function, and $\Omega$ is the regularization term that ensures the retention of knowledge across domains.

\textit{Example:} The illustration in Figure \ref{fig:cl_type} (middle) shows the domain incremental learning concept, specifically in the medical field. The model is initially trained to distinguish between healthy and tumor tissue from a specific data center. As new data from a different data center with varying imaging quality, contrast, or patient demographics becomes available, the model should be able to adapt to this new domain. This robust adaptation process allows the model to generalize its segmentation abilities across diverse acquisition conditions without requiring any task-specific identifiers. 

\subsubsection{Class-Incremental Learning}
\label{CIL}

\textit{Theory:} Class-incremental learning is claimed to be the most challenging type of CL as it should be able to handle all the tasks at once. In this protocol, the model should continually learn to differentiate between an increasing number of classes over time and evaluate all observed classes at test time, which is essential in the medical field.

\textit{Formulation:} Given a growing sets of classes $C_1$ through $C_n$, the aim is to learn mapping from $X_i$ to $Y_i$ for each class. The main objective is to optimize the model parameters $\theta$ to incorporate the new classes $C_c$ and keep the existing classes distinctiveness: 

\begin{equation}
    \arg \min_{\theta_{\zeta+1}} \left[ \mathcal{L}\left(f\left(X_{C_i}, \theta_{\zeta+1}\right), Y_{C_i}\right) + \lambda \Omega\left(\theta_{\zeta+1}, \theta_{\zeta}\right) + \beta \Psi\left(\theta_{\zeta+1}\right) \right]
\end{equation}

\noindent where $\Omega$ is a class-specific regularization function that tries to retain the knowledge of previously learned classes; here $\theta_{\zeta}$ are parameters of previously learned classes. The term $\Psi$ is also a regularization term that aids in maintaining distinct class representation by preventing overlap between the feature representation of old and new classes. Note that $\zeta+1$ does not mean adding one class but adding any number of classes.

\textit{Example:} Figure \ref{fig:cl_type} (right) depicts an example of class-incremental learning in the medical domain. Initially, the model is trained to differentiate between background and brain tumors. As the new data arrives, the model learns to identify the new class of lung tumors without compromising its ability to recognize previously learned classes. This is achieved by maintaining separate and distinct feature representations for each class. Also, the model can differentiate between all the classes - background (0), brain tumor (1), and lung tumor (2) without any task identifier. 

\textit{\textbf{Remark}}: DIL is applicable when data arrives from different centers or modalities, as shown in Figure \ref{fig:cl_full} (Centers). In contrast, CIL is relevant when introducing a new set of classes, such as adding a lung tumor in an existing setting of diagnosing brain tumor, as shown in Figure \ref{fig:cl_full} (Predictions). TIL is a sequential multi-task type where the model is incrementally given a new task, and the task is a clear separation from the previous ones. For example, adding a lung tumor in an existing setting of diagnosing brain tumor and expecting the model to output any or both the tasks based on the input requirement. 

\subsection{Evaluation Metrics}

Several evaluation metrics are commonly used in continual learning. These metrics help assess performance across tasks in an incremental learning setting. Below, we summarize some of the most frequently used metrics:

Let $A_j^i$ (where $i, j = 1, 2, \dots, T$) represent the accuracy of a model trained after task $j$ and tested on the data from task $i$. Here, $A_j^i$ refers to the accuracy of the model trained after task $j$ and tested on all previously seen tasks (test data from tasks $0, 1, \dots, j$).

The arrows $\uparrow$ and $\downarrow$ denote whether "higher is better" or "lower is better," respectively.

\begin{itemize}
    \item \textbf{Average Accuracy (AvgACC)}: $\uparrow$
    \[
    \text{AvgACC} = \frac{1}{T} \sum_{i=1}^{T} A^i
    \]
    
    \item \textbf{Final Average Accuracy (FAA)}: $\uparrow$
    \[
    \text{FAA} = \frac{1}{T} \sum_{i=1}^{T} A_T^i
    \]
    
    \item \textbf{Backward Transfer (BWT)}: $\uparrow$
    \[
    \text{BWT} = \frac{1}{T-1} \sum_{i=2}^{T} \sum_{j=1}^{i-1} \left( A_j^i - A_i^i \right)
    \]
    
    \item \textbf{Average Forgetting (AvgF)}: $\downarrow$
    \[
    \text{AvgF} = \frac{1}{T-1} \sum_{i=1}^{T-1} \left( \max_{k=1, \dots, T-1} A_i^k - A_T^i \right)
    \]
\end{itemize}

\section{Taxonomy}
\label{taxonomy}

\begin{figure}[b]
\centering
\includegraphics[width=\linewidth]{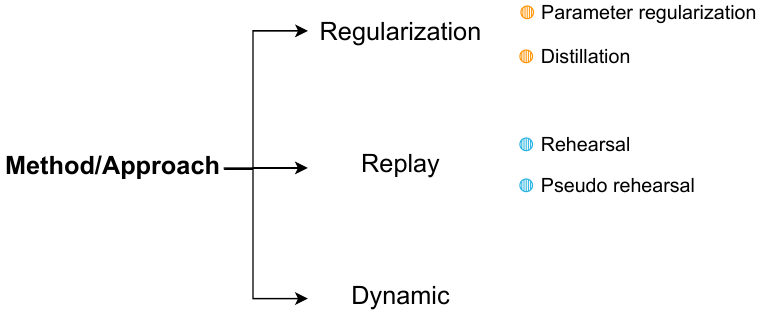}
\caption{Taxonomy of Continual Learning Approaches.}
\label{taxonomy_fig}
\end{figure}

Numerous studies have addressed incremental learning in medical images in recent years. We organize these methods taxonomically from the approach they take to implement, that is, regularization-based, replay-based, and dynamic model-based. Replay-based methods concentrate on solving the CL with exemplars, which can be further classified into data memory and feature replay. Regularization-based methods regularize the model parameters from drifting away. Finally, dynamic model methods either isolate parameters based on the task or expand the model structure to accommodate more task knowledge. The following sections discuss CL methods using these three approaches.

\subsection{Regularization Approaches}

\begin{figure}[t]
\centering
\includegraphics[width=\linewidth]{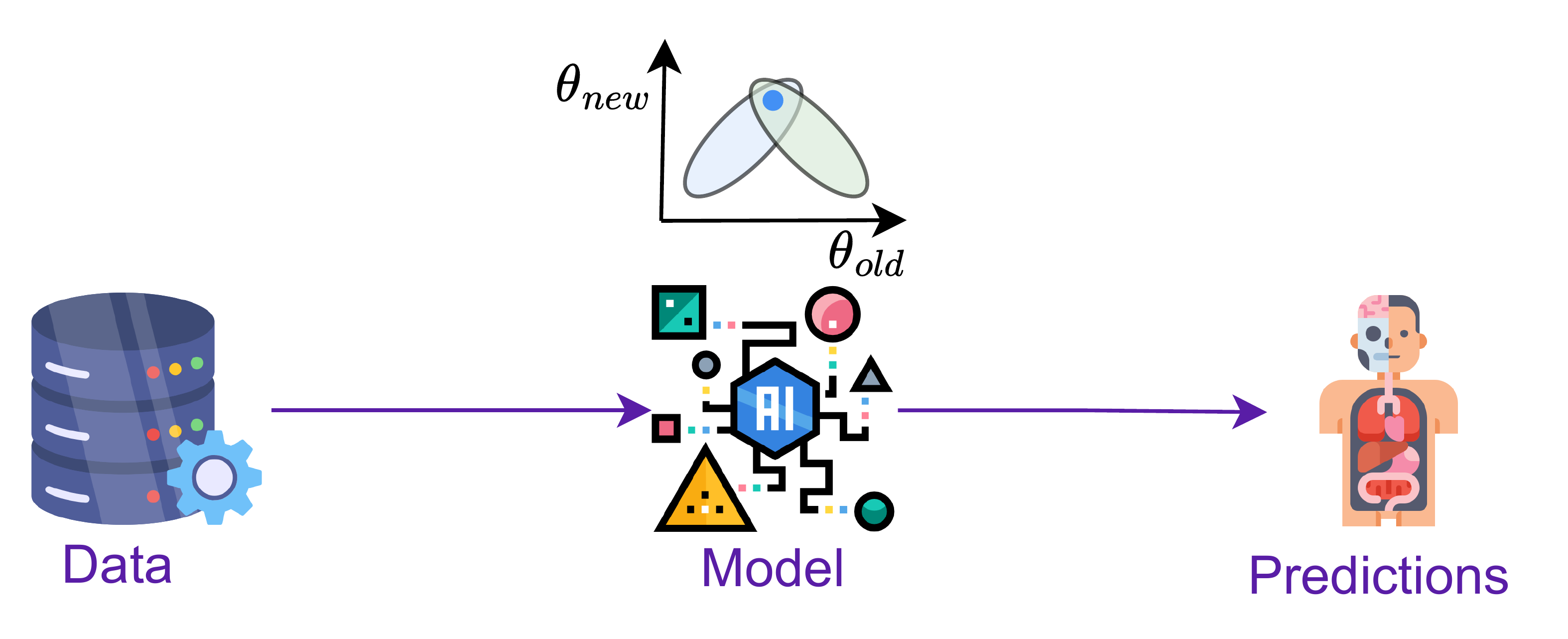}
\caption{Regularization Based Approach: The model is updated by constraining the parameter changes. Assuming $\theta_{new}$ and $\theta_{old}$ are the parameters of the updated and previous models, respectively. The objective is to find an optimal point between the parameter space of these models.}
\label{regularization}
\end{figure}

Regularization-based continual learning offers an effective solution for learning new tasks while preventing catastrophic forgetting of previous ones. The key concept behind this approach is to minimize the drift in the learned feature or parameter space of the model, as shown in Figure \ref{regularization}. By doing so, the model can learn new tasks without inducing significant changes in the parameters that are useful for old tasks, enabling it to retain prior knowledge \citep{wang2023comprehensive}. Regularization-based CL incorporates additional loss terms to prevent model drift between old and new tasks. This is typically achieved by maintaining a copy of the model from the previous task or by preserving the model predictions. In medical applications, regularization-based CL are valuable because they do not require access to stored samples, which is a challenge because of privacy concerns. In addition, it does not dynamically increase network size. This makes regularization-based CL models lightweight and versatile, making them particularly suitable for various medical settings. Regularization-based methods fall into two primary categories: parameter regularization and functional regularization, depending on whether they focus on the network parameters or output.

\subsubsection{Parameter Regularization}
Parameter Regularization focuses on the change in network parameters when learning new tasks. This is achieved by introducing an additional loss term to penalize alterations in the parameters deemed important for the previous tasks. Typically, this involves maintaining a frozen copy of the old model to calculate the change in network parameters with the introduction of new tasks. Several studies have demonstrated the effectiveness of this approach in the medical field. \cite{baweja2018towards} incorporated a quadratic penalty in the loss function to address parameter changes important for prior tasks. It follows the Elastic Weight Consolidation (EWC) method using the Fisher information matrix to quantify each parameter's contribution \citep{kirkpatrick2017overcoming}. Their work comprehensively analyzed the effectiveness of the EWC method in medical imaging applications. Similarly, \cite{van2019towards} applied a strategy of penalizing changes in important parameters for brain MRI segmentation, demonstrating its efficacy in transfer learning by pre-training on a high-quality dataset before fine-tuning it on a smaller, lower-quality dataset. This provides an insight into the application of regularization-based CL methods in transfer learning for low-quality datasets, which are common in the medical domain. Another study by \cite{lenga2020continual} used EWC and Learning Without Forgetting (LWF) methods for multidomain chest X-ray classification and compared their performance against joint training. Their work showed that the LWF outperformed the EWC approach and achieved a backward transfer on par with joint training on the new domain conducted with 60-80\% of the data replay of the previous domain. \cite{chen2022breast} combined the hybrid-averaging operation and EWC into CNN models for breast cancer image classification tasks and found that the ResNet \citep{he2016deep} outperforms AlexNet \citep{krizhevsky2012imagenet} and DenseNet \citep{huang2017densely} in terms of average accuracy and average forgetting. In addition, \cite{derakhshani2022lifelonger} and \cite{quarta2022continual} offered a benchmark for various regularization-based CL methods, including EWC \citep{kirkpatrick2017overcoming}, LwF \citep{li2017learning}, iCaRL \citep{rebuffi2017icarl}, MAS \citep{aljundi2018memory}, and EEIL \citep{castro2018end}, and showed that iCaRL and EEIL provide superior performance compared to other methods.

More recently, \cite{ozgun2020importance} proposed a modified version of the memory-aware synapses (MAS) \citep{aljundi2018memory} method by imposing learning rate (LR) constraints on each parameter, instead of direct parameter regularization. In this approach, each parameter or an entire CNN kernel is assigned a specific LR based on its importance for the previous tasks. The authors also proposed freezing the important parameters and fine-tuning the unimportant parameters using LR constraints. Furthermore, \cite{chen2021targeted} introduced the Targeted Gradient Descent (TGD) method to fine-tune a model without forgetting previous knowledge by leveraging the Kernel Sparsity and Entropy (KSE) metric in a positron emission tomography (PET) reconstruction study. TGD provides an effective method of reusing redundant kernels in a pre-trained CNN for learning new tasks while preserving the balance between the rigidity and plasticity of the models. Another study by \cite{bayasi2021culprit} used a pruning criterion that allows a fixed network to learn new data domains sequentially over time by detecting the \textit{culprit} units associated with the wrong classification. These network units are dedicated to the subsequent learning of new domains. MUSCLE \citep{liao2022muscle} proposed the addition of model parameter constraints for pretraining a robust backbone in the classification, segmentation, and detection of X-ray images. It utilizes momentum updates between model parameters to prevent catastrophic forgetting during pretraining when moving from one task to another. \cite{ranem2022continual} combined a ViT \citep{dosovitskiy2020image} and nnUNet \citep{isensee2021nnu} models for continual hippocampus segmentation using the EWC and PLOP methods\citep{douillard2021plop}. Their proposed model utilizes the self-attention mechanism of transformers to maintain knowledge and consequently alleviate catastrophic forgetting, in contrast to the purely convolutional UNet. Nevertheless, interestingly, their experiments showed that applying regularization to the transformer component over time has an adverse effect on knowledge retention, as regularization disrupts the self-attention mechanism. In contrast, it is preferable to discourage substantial changes in the convolutional layers to retain prior knowledge without compromising the plasticity. Furthermore, \cite{zhang2021comprehensive,zhang2023s} proposed a comprehensive importance (CI) method for selecting important parameters based on their contribution to the shape and certainty of output segmentation. They demonstrated that maintaining such parameters can preserve the important shape and semantic information of the segmentation task in CL settings. In another study by \cite{shu2022replay}, regularization in the gradient space of weights was implemented. This involved updating the model weights using gradients orthogonal to the important parameters for the previous task(s). To further enhance the performance, they incorporated data replay into their methods. Furthermore, \cite{wang2022incremental} introduced a technique to mitigate catastrophic forgetting in lung nodule detection studies. Their approach was based on a modified EWC and incorporated feature distillation for enhanced performance.  Conversely, \cite{liu2023incremental} used a frozen copy of the model along with other techniques to adapt an off-the-shelf segmentation model for different anatomical regions. The two model copies were combined by taking a weighted average of the corresponding layers at the end of each training round, allowing for the retention of previous knowledge while acquiring new task-specific information.

\subsubsection{Functional Regularization}
Functional Regularization impedes the intermediate or final prediction of the model by using a distillation-based approach, where the old model acts as a teacher to the new model. This approach does not include any explicit constraints on the model parameters and only uses the model's output to keep the learned feature space close between the tasks.

\cite{kim2018keep} proposed learning an inverse function that maps the model's logits to the feature space and uses it to regularize the feature space of the model in the next task. They demonstrated the effectiveness of their method in a multidomain, single-task X-ray classification scenario. An interesting domain incremental study provided a dual-distillation method combined with a replay to mitigate forgetting \citep{li2020continual}. Dual-distillation is carried out using the logits from the old and a new fully fine-tuned model to train an updated model that performs well on new and old tasks. Recent work on manifold learning \cite{akundi2021manifold} used a reformer network to augment any input domain image to lie in the same data manifold. This, combined with various regularization-based methods, was shown to be effective in reducing forgetting due to domain shifts in the clinical data. Moreover, a multimodal study by \cite{patra2021multimodal} on ultrasound and eye-tracking data utilized distillation to train a model on a continual stream of data, whereas \cite{patra2021multimodal} proposed an architecture that leverages the simultaneous availability of two or more datasets to learn a disentanglement between the content and domain in an adversarial manner. This allows a domain-invariant content representation that lays the basis for the continual semantic segmentation of MRI images.\cite{he2021incremental} use a probability-map alignment scheme to integrate the previous and current map for a segmentation task. They further used knowledge distillation to transfer the knowledge from the integrated probability map to the current model, thereby improving the segmentation performance of the current model. \cite{chakraborti2021contrastive} presents a semi-supervised continual learning method combined with a contrastive loss. Their approach was to first train an auto-encoder with a small amount of labeled data in the usual supervised manner, followed by an unsupervised adaptive regularization-based contrastive learning approach.

Another work by \cite{liu2022learning} introduced a method for multi-organ segmentation in CT images. Their methodology uses lightweight memory modules and a tailored regularization loss function, allowing the model to effectively utilize replayed examples to prevent forgetting. Here, the introduced loss function constrains the learned feature space to maintain the feature of an old class close to its mean representation within the memory module, while ensuring that the features of a new class are distinct from all representations of the old class. Furthermore, \cite{tian2022multi} proposed a multi-scale multitask CL framework. Their method combined contrastive learning with distillation-based regularization and data rehearsal to enhance learning without catastrophic forgetting. \cite{li2022domain} used feature whitening to regularize the important domain in-variant features in a domain incremental cardiac image segmentation study. This feature regularization was augmented by a generative feature replay, where they proposed a base conditional generator along with a style module to produce images for replay. Furthermore, \cite{roy2023l3dmc} introduced the use of a mixed-curvature embedding space for knowledge distillation to prevent catastrophic forgetting. The authors employed a combination of Euclidean and hyperbolic spaces to project the learned feature embeddings, emphasizing their importance for complex data, especially in medical imaging. Another recent study by \cite{gao2023incremental} used a Kullback–Leibler (KL) divergence-based loss function to preserve the models’ knowledge by penalizing the change in output of the new and old models. More recently, the Fourier test–time adaptation (FTTA) method was presented for classifying medical images with domain gaps \citep{huang2023fourier}. This technique utilizes a Fourier-based domain adjustment strategy along with a multilevel consistency approach to regularize the model's output. The proposed method is based on a multilevel consistency measurement designed for self-correcting predictions.


Overall, regularization-based methods offer an efficient and privately protected approach to addressing catastrophic forgetting in CL. These methods do not introduce significant computational overhead or privacy concerns, making them initially well-suited for real-life medical applications. However, regularization-based methods also have notable limitations. Empirical evidence shows that these methods often lead to significant performance deterioration over time, particularly when learning diverse and distinct tasks. The trade-off between plasticity and stability can result in models that become increasingly rigid or too plastic, ultimately limiting their practical use. Additionally, as more tasks are learned, the gap between the performance of regularization-based models and the ideal of maintaining individual models grows wider. These challenges make regularization-based methods less effective for long-term use in dynamic environments. Recent efforts to combine regularization-based approaches with other CL techniques have shown promise in mitigating these issues, which could be a key direction for making regularization-based CL methods more practical in medical applications.

\subsection{Replay-Based Approaches}

\begin{figure}[t]
\centering
\includegraphics[width=\linewidth]{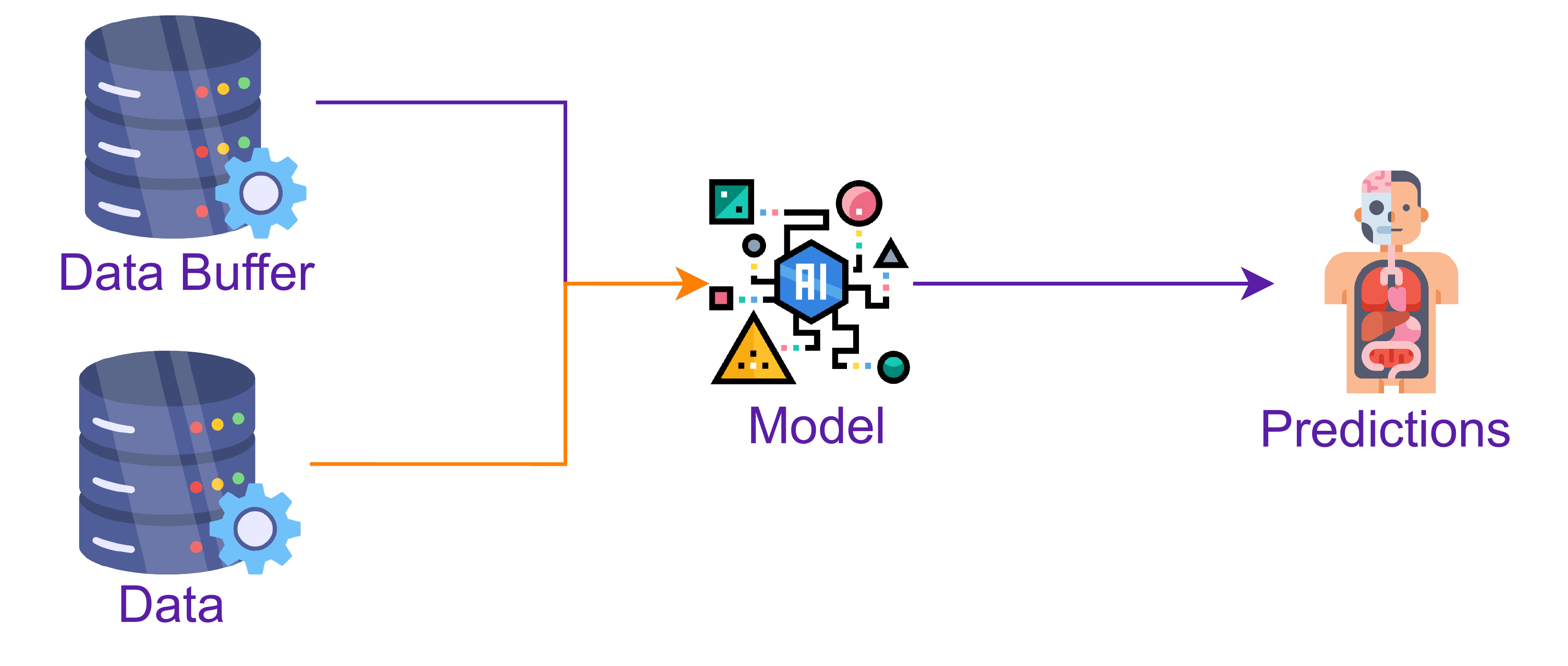}
\caption{Replay Based Approach: Store some previous samples as a replay buffer to preserve what has been learned.}
\label{replay}
\end{figure}

Replay-based methods in CL were designed to address the problem of catastrophic forgetting in sequential learning scenarios. These methods use a replay mechanism or memory buffer to preserve and reintroduce prior experiences while training new tasks to reduce forgetting, as depicted in Figure \ref{replay}. This buffer typically contains samples, such as input-output pairs and representations of past data,  preserving knowledge that the model has learned previously.

The ability of replay-based techniques to retain learned knowledge while adjusting to new information or tasks is a crucial feature that makes them highly advantageous and applicable to the medical domain. This ability to retain information is critical, as it ensures that the model retains the critical patterns, diagnoses, or treatment protocols that it has learned from previous experiences. Replay-based models facilitate the exchange of knowledge and skills among various healthcare facilities and regions. These strategies can also help with privacy and data constraints by effectively utilizing a smaller collection of experiences. This reduces the need to constantly access or store new patient data, while maintaining the confidentiality of sensitive information.

\cite{venkataramani2019towards} use Memory-augmented neural networks(MANN) for continuous domain adaptation for the task of semantic segmentation. They proposed a single framework for domain adaptation in which images similar to the target image and their respective textual and shape features are fetched using wavelet features of the target image as a query. These context features, along with the target image features, were embedded into the model trained using only the source domain. \cite{ravishankar2019feature} proposed a single approach for the different variants of lifelong learning, such as multi-task setting and single incremental task setting, aiming to ensure data privacy and mitigate catastrophic forgetting. They employed a pseudo-rehearsal strategy using a finite memory module and ensured class separation using a composite loss. Furthermore, they introduce feature transformers to learn new representations from the old data as well as the new data. \cite{hofmanninger2020dynamic} proposed a rehearsal method that infers the data shifts by adopting a dynamic memory module. Their memory update strategy was designed based on a fixed set of rules, and the features were replaced with a high-level metric calculated using gram matrices. \cite{perkonigg2021dynamic} employs the same strategy but also introduces an additional module called the pseudo-domain (PD) model, which detects clusters of similar styles from the continuous stream of data. These pseudo-domains are seen as proxies for the unknown, real domains, and are used to balance the memory and training process. \cite{srivastava2021continual} also proposed another domain incremental learning that leverages vector quantization approach under limited memory constraints to effectively store and replay hidden representations where no clear domain-shift boundaries exist and no a-priori knowledge is possible. \cite{wang2021lifelong} introduced a method that integrates medical expertise into disease diagnosis by considering context and medical entity characteristics in order to transfer knowledge to new stages. Their work was among the pioneering studies exploring CL within medical domains, specifically focusing on diagnosing diseases based on clinical notes. Additionally, they introduced a new technique called embedding episodic memory and consolidation (E2MC) to avoid catastrophic forgetting in disease diagnosis tasks. \cite{morgado2021incremental} delve into continual learning within dermatology, employing four distinct methods: naive, elastic weight consolidation, averaged gradient episodic memory, and experience replay. These methodologies allow adaptation to new conditions while safeguarding previously acquired knowledge in the field.

To overcome the problem of insufficient sensitive patient data for digital biomarker model training owing to privacy laws,  \cite{shevchyk2022privacy} proposed a Privacy-Preserving Synthetic Data Generation pipeline that uses a GAN to generate anonymous medical samples. Privacy is ensured using a Siamese Neural Network (SNN) with triplet loss to evaluate the generated data. The effectiveness of this method was demonstrated using a Respiratory Sound Classifier that collects respiratory symptom information and is relevant for conditions such as Asthma, COPD, or COVID-19.

Similarly, \cite{liu2022new} pioneered a generative replay-based approach in semantic segmentation tasks for medical imaging by introducing SegGAN to prevent catastrophic forgetting. This GAN generates both previous images and their corresponding pixel-level labels, addressing privacy limitations while preserving the acquired knowledge for the subsequent learning stages. Additionally, they proposed a unique filtering mechanism that ensures the selection of high-quality generated data by maintaining consistency between the pseudo-labeling and generative replay techniques. \cite{li2022continual} propose the use of multiple generators to improve the quality of replayed samples. Instead of using generators to generate higher-resolution images, this study uses a fixed feature extractor and focuses on feature generation, which requires fewer parameters and fewer computations. \cite{li2022class} introduced a fresh approach for CL in gesture recognition, specifically for adding new gesture classes. This method focuses on managing out-of-distribution (OOD) memory and improving rehearsal to consistently learn new gestures. Additionally, it incorporates an energy-based loss to penalize out-of-distribution gesture examples and further refine the model during training. \cite{ayromlou2022class} introduces an innovative data-free framework for incremental class learning.  This approach initially creates synthetic data from the model trained on past classes to form a class impression, and then updates the model by merging the synthesized data with new class data. They employ various loss functions, such as cosine normalized cross-entropy, margin loss, and intra-domain contrastive loss, to tackle class imbalances, differentiate between previous and new classes, and ensure the model's generalization from synthetic to real data. \cite{zhang2022learning} introduce a new method, SMG (Synchronous Memorizability and Generalizability) -learning, to enhance both memorization and generalizability. It presents the Synchronous Gradient Alignment (SGA) objective of improving network memorization and adaptability across previous and new sites. To streamline optimization without excessive computational load, the dual-meta algorithm was developed. Moreover, to enhance rehearsal effectiveness, the replay buffer is modified to include diverse data from various sites, thus minimizing redundancy.

\cite{bera2023memory} introduce a novel solution to improve image selection to create the memory bank by ranking and selecting images based on their contribution to the learning process. In situations such as medical applications where storing previous data is limited by privacy regulations,  \cite{chen2023generative} presented Generative Appearance Replay for continual Domain Adaptation (GarDA), an approach employing generative replay techniques. GarDA facilitates the gradual adjustment of a segmentation model to new domains by utilizing unlabeled data. Similarly, \cite{thandiackal2023multiscale}  proposed a generative feature-driven image replay with a dual-purpose discriminator to align features for continual learning in unlabeled domains, achieving state-of-the-art results in histopathology datasets. Unlike single-step unsupervised domain adaptation (UDA), continual adaptation across various domains enables the utilization and consolidation of knowledge from different domains throughout consecutive learning phases. \cite{10341107} proposes a data selection strategy using Variational autoencoders (VAE) and adversial network. Through this approach, they retain training efficiency by always selecting a fixed subset size from the entire dataset. \cite{kim2024continual} proposes a continual learning for multicenter studies, eliminating the need for a central server and mitigating the risk of catastrophic forgetting of previously acquired knowledge. The framework involves a method selection process for continual learning, acknowledging that no single method suits all datasets and utilizes fake data generated by a generative adversarial network to evaluate the methods.


Replay-based techniques can be applied in medical settings; however, they bring additional challenges related to model interpretability, regulatory validation, and ethical concerns, all of which must be addressed to ensure the safety and effectiveness of AI-driven healthcare solutions. The primary issue with rehearsal-based methods is the consideration of data privacy, as storing subsets of data may not be feasible due to privacy regulations. This challenge can be mitigated through pseudo-rehearsal or generative rehearsal techniques, where synthetic samples are generated instead of storing actual data. Overcoming these challenges is essential for fully leveraging the potential of continual learning techniques in healthcare while safeguarding patient privacy and safety.

\subsection{Dynamic Model Approaches}

\begin{figure}[t]
\centering
\includegraphics[width=\linewidth]{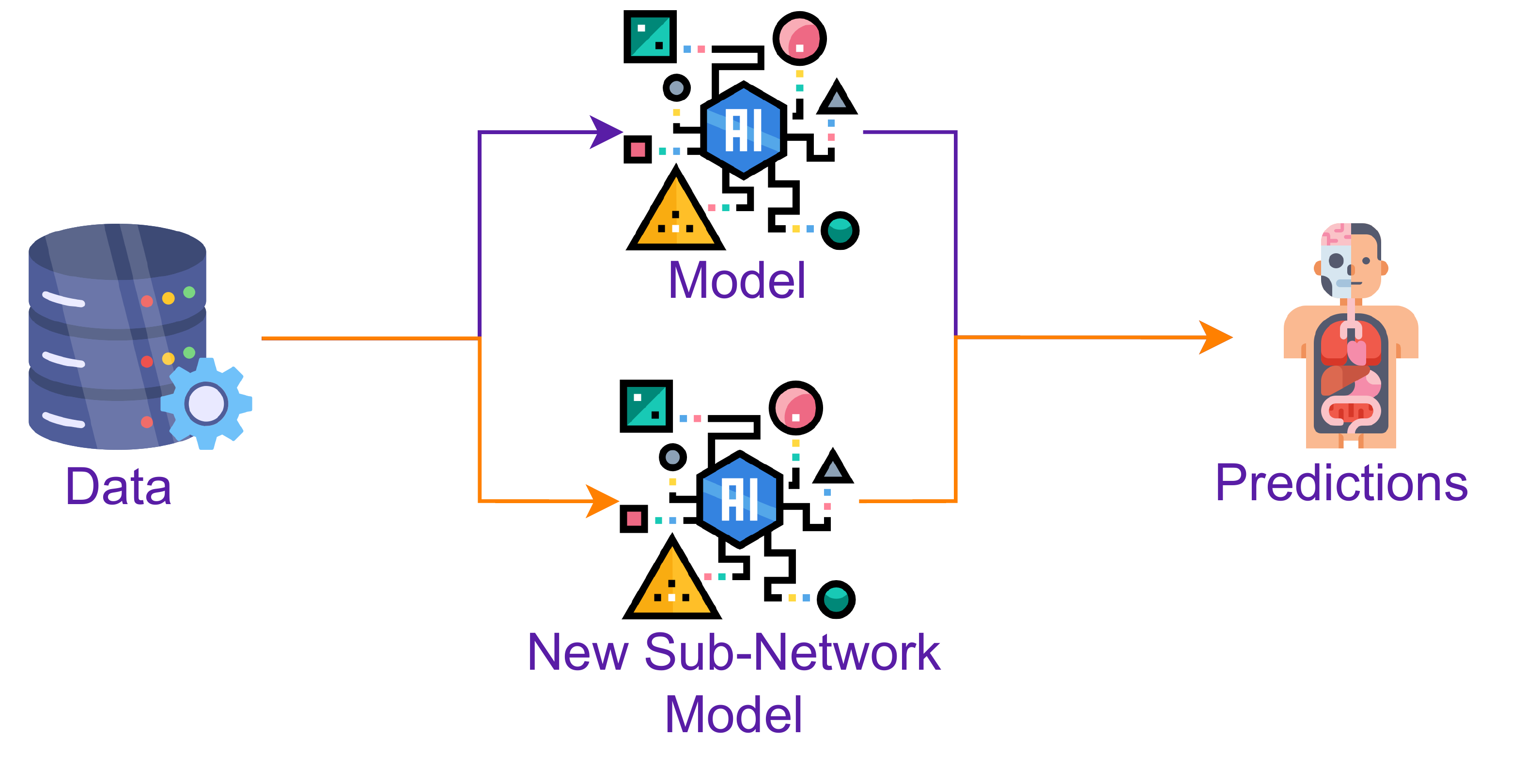}
\caption{Dynamic Based Approach: allocate new parameters for each of the new tasks by adding a new sub-network for each task.}
\label{dynamic}
\end{figure}

Previous methods were either limited by the model's parameter change in subsequent tasks or relied on rehearsal techniques. A more recent and promising solution involves intelligent introduction of new parameters for each task (Figure \ref{dynamic}). In other words, shared features are maintained among tasks to avoid repeated training, similar to ensembling. Moreover, employing a set of new parameters enables more effective learning of task-specific features.

Dynamic models offer a direct solution by allocating specific segments of the model to distinct tasks. A study by \cite{karani2018lifelong} addressed domain shifts in lifelong learning scenarios using domain-specific batch normalization layers for MRI segmentation. Variance in MRI image intensities, caused by multiple factors, leads to this domain shift \citep{preboske2006common,jovicich2006reliability}.

However, another challenge arises when we lack knowledge about the task or which sub-network to choose during inference for particular data. For example, if a model is trained on images from three different scanners, determining the domain during inference requires image metadata to activate the corresponding model segment. \cite{gonzalez2020wrong} tackled this issue by training separate UNet models for domain segmentation. To discern the domain during inference, they utilized separate Variational Autoencoders (VAEs) for each domain and selected the domain with the minimal VAE reconstruction loss. Another study by \cite{gonzalez2022task} introduces a task-agnostic continual learning approach applicable across diverse deep learning architectures. Their method involved utilizing the Mahalanobis distance to identify domain changes. Upon detecting such changes, they expand the network, ensuring a nonlinear expansion. During inference, the network leverages this distance measurement to determine task ID, thereby facilitating efficient task identification. Similarly, \cite{bayasi2023continual} selected a network based on a distance-based metric for each test dataset. In contrast, \cite{zhao2023class} used different prompts for each task and used a matching strategy to choose the correct prompt at inference. 

Recently, advanced techniques have been applied to address the challenges of multiorgan segmentation (\cite{gonzalez2023lifelong}). These methods operate in a class incremental setting, requiring the model to segment all the regions observed in previous tasks during inference. \cite{zhang2023continual} highlights the significance of dynamically expanding models to encompass new data and classes for segmenting multiple organs and tumors. Their architecture features lightweight class-specific heads that replace traditional output layers, enabling separate predictions for both new and previously learned classes. Leveraging extensive image-text co-training, the authors embedded contrastive language-image pretraining (CLIP) embeddings into organ-specific heads to capture each class's semantic essence. Research by \cite{you2022incremental} introduced an innovative incremental transfer learning framework designed to progressively address multi-site medical image segmentation tasks. They proposed strategies at both the model and site levels aimed at incremental training to enhance segmentation, improve generalization, and boost transfer performance, particularly in settings constrained by limited clinical resources.

\cite{ji2023continual} introduced an innovative architectural framework for semantic segmentation. Their approach learns a unified deep segmentation model capable of segmenting 143 whole-body organs by training a single unified deep segmentation model. For each segmentation task, a trained encoder remained fixed, whereas a new decoder was gradually added. To handle model complexity, they utilized a progressive pruning strategy involving neural architecture search and teacher-student knowledge distillation. This strategy helps maintain performance while controlling the size of the model.

The practice of dynamically adjusting models extends to classification tasks, as observed in a recent study by \cite{zhang2023adapter}  on various medical imaging classification tasks. Building upon insights from \cite{ding2022delta,houlsby2019parameter}, they proposed incorporating a lightweight, adaptable module, called an adapter, into a pre-trained and fixed encoder. This addition enables the efficient continual learning of new knowledge by adapting the model for each task to grasp a new set of classes. To address the task identification challenge in continual learning scenarios, task-specific heads were introduced. These heads include an extra class that assimilates all previously learned old classes. Another classification study \citep{chee2023leveraging} introduced a dynamic architecture featuring expanding representations. Their approach involved incorporating a high-level feature extractor for each task while maintaining a shared low-level feature extractor. This architecture aims to retain common features while learning task-specific characteristics incrementally, asserting the improved preservation of information during each step. In addition, another work by \cite{mousser2022idt} proposed an Incremental Deep Tree (IDT) framework tailored for biological image classification. The IDT framework adopted a hierarchical tree-like structure, constructing a new model for each new class by appending branches linked to previously learned tasks. These branches serve to regulate previously acquired knowledge by leveraging replay data and mitigating the issue of catastrophic forgetting. As a result, the framework continuously updates knowledge, enabling the root model to predict new classes while retaining past information and ensuring consistent accuracies for each learned class. \cite{xie2023task} proposes a simple yet effective approach by training batch normalization layer for every task while keeping the rest frozen.

However, while dynamic models offer a logical solution to the problem, identifying the task ID during inference is an open research area for various tasks such as classification and multiclass segmentation. Although dynamic models have proven effective in multi-organ segmentation, this approach may not be optimal, as it requires extensive training and might not be practical. Mitigating forgetting is not the only objective of continual learning.

\section{Discussion}
\label{discussion}

\begin{figure*}[ht]
\centering
\includegraphics[width=\linewidth]{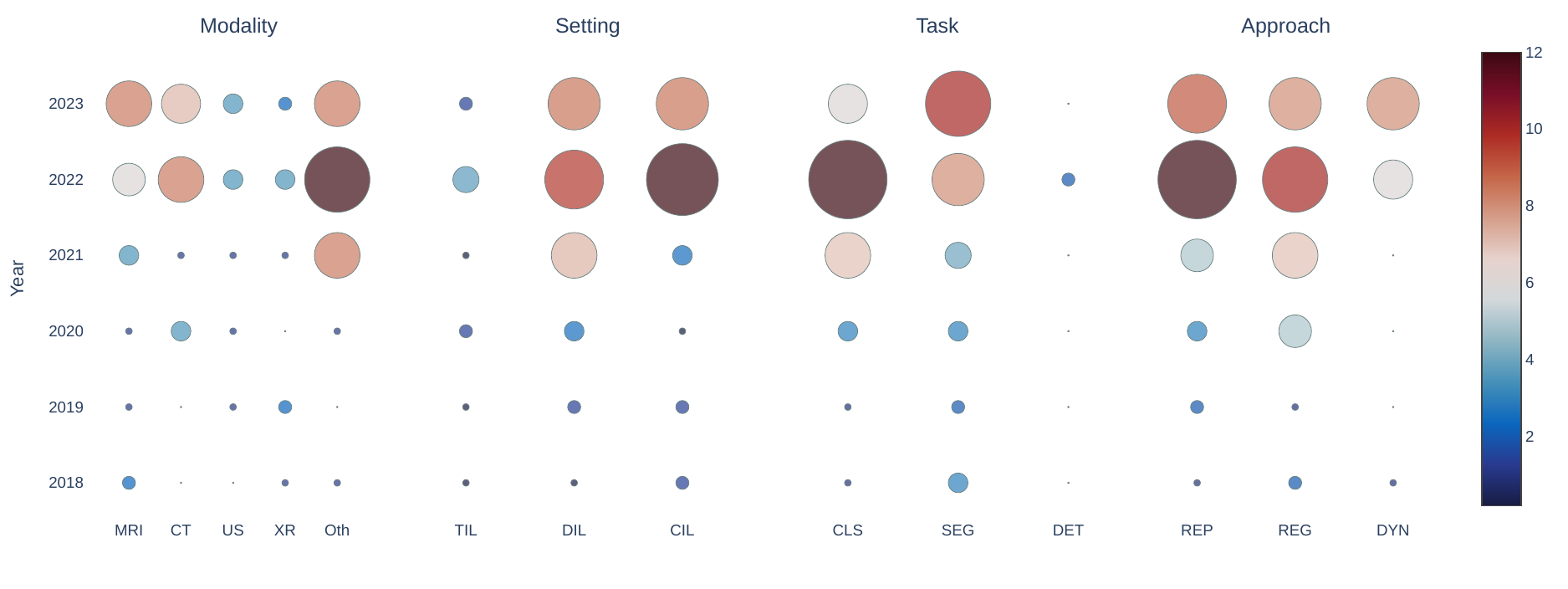}
\caption{Trend from various perspectives in CL in medical. (a) Modality. MRI: Magnetic Resonance Imaging, CT: Computed Tomography, US: Ultrasound, XR: X-Ray, Oth: Others, (b) Setting. TIL: Task Incremental Learning, DIL: Domain Incremental Learning, CIL: Class Incremental Learning, (c) Task. CLS: Classification, SEG: Segmentation, DET: Detection, (d) Approach. REP: Replay, REG: Regularization, DYN: Dynamic.}
\label{fig:bar}
\end{figure*}

Within the medical domain, notable progress has been made in the application of the CL methods. However, the direct use of methods from natural image domains to medical data presents several challenges. Hence, various methodologies have been proposed to address these challenges. 

In most studies, there has been a surge in the use of the CL method in medical settings. An ideal and resilient CL model should possess three key characteristics for feasible deployment: maintaining stability, adaptability, and operating within resource constraints. Initially, while training on the first task, the model should be sufficiently flexible to adjust and accommodate new data that may arise in the future. In addition, it is essential for models to retain the knowledge acquired from previous tasks while accommodating new learning tasks. Furthermore, considerations of the size and complexity of the model are important, particularly in deployment scenarios. Therefore, there is a critical need to maintain the model efficiency and scalability to ensure practical feasibility in real-world applications. Moreover, there is a recognized need to develop task-agnostic solutions that are deemed more practical and applicable to real-world scenarios. Consequently, a significant portion of research efforts has been directed towards the development and evaluation of such methodologies. 

In this section, we provide a comprehensive overview of the reviewed papers, particularly focusing on CL methods within the medical domain. We identified recent trends, highlighted some of the challenges associated with practical settings, and suggested future directions.

\subsection{Trends}
\label{trend}

In the medical field, the primary focus of tasks centers around segmentation and classification, with relatively less emphasis on detection. The presented figure \ref{fig:bar} illustrates a recent growth in both segmentation and classification tasks, with nearly equal attention directed towards both. Segmentation models demand extensive training and computational resources. 

Consequently, most of the research has concentrated on regularization approaches to address continual learning challenges. Over half of the reviewed works opted for regularization-based methods to tackle this issue effectively. Furthermore, there has been a recent surge in dynamic model approaches, which, although are less memory efficient and complex to train, offer a simpler solution \citep{van2022three}. Dynamic models, which involve maintaining an ensemble of models for various tasks, offer a straightforward means to handle the problem, contrasting with alternative approaches that may entail intricate learning processes.

\begin{figure}[ht]
\centering
\includegraphics[width=\linewidth]{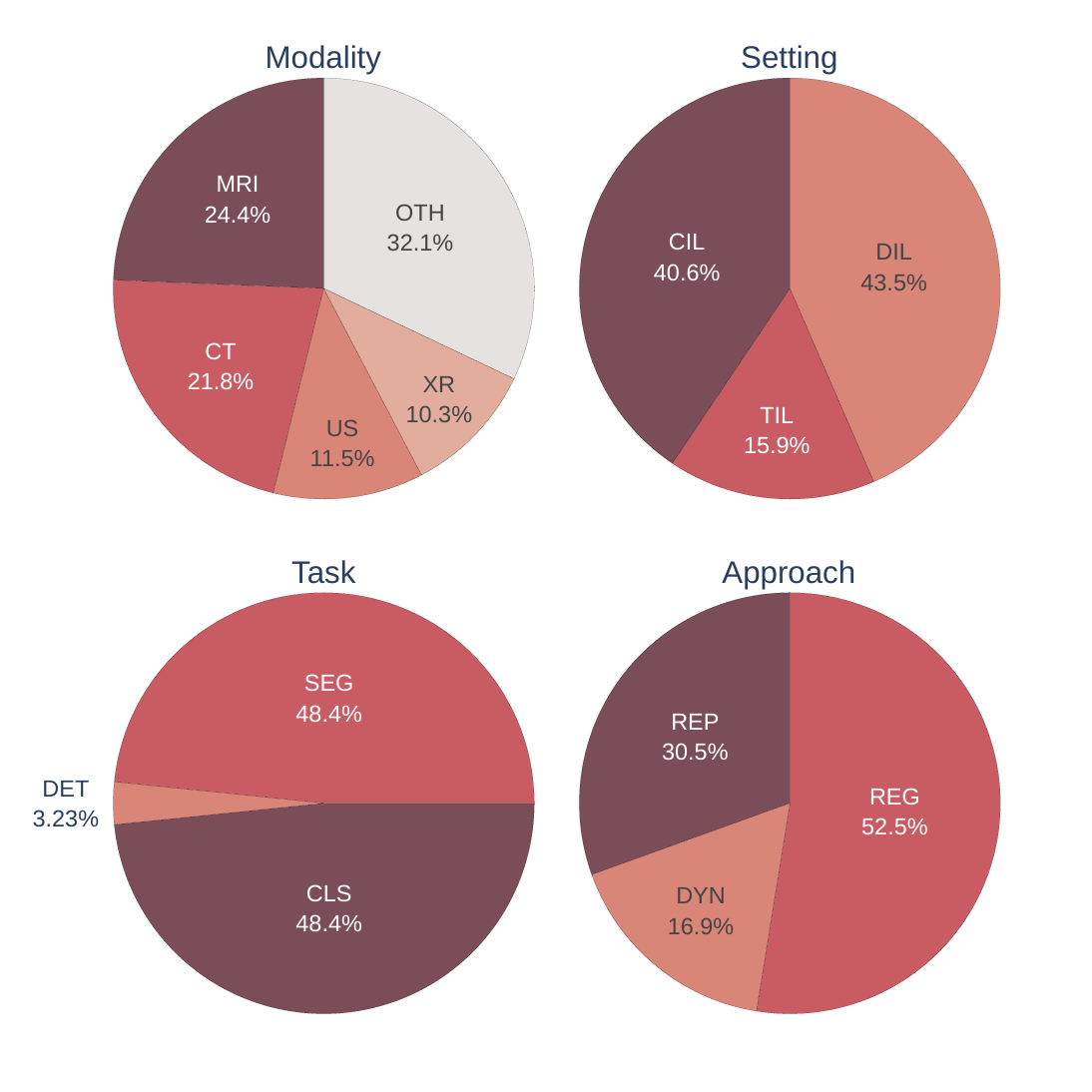}
\caption{Distribution of studies from various perspectives in CL in medical. (a) Modality. MRI: Magnetic Resonance Imaging, CT: Computed Tomography, US: Ultrasound, XR: X-Ray, Oth: Others, (b) Setting. TIL: Task Incremental Learning, DIL: Domain Incremental Learning, CIL: Class Incremental Learning, (c) Task. CLS: Classification, SEG: Segmentation, DET: Detection, (d) Approach. REP: Replay, REG: Regularization, DYN: Dynamic.}
\label{fig:pie}
\end{figure}

 Among the different incremental learning scenarios, class-incremental learning stands out as the most challenging and practical setting, where, during inference, there is no specific knowledge available regarding the task of data. Figure \ref{fig:pie} demonstrates the significant attention given to the task-agnostic approaches, which is noteworthy. Additionally, most of the literature in this field concentrates on developing methods suitable for task-agnostic settings. 

Reproducibility in scientific research is important and is often achieved through the sharing of code and the utilization of publicly available datasets. However, challenges to reproducibility persist, notably when researchers omit details about data splits or utilize private data, thereby hindering accurate replication. Although recent studies commonly utilize publicly available data, many have not shared their code, which significantly hampers reproducibility. Furthermore, most recent research works compare their findings with previous state-of-the-art approaches, providing valuable insights into the field's progress and establishing benchmarks for future studies. Overall, addressing challenges such as data split information, code sharing, and comparison with previous works is crucial for enhancing the reproducibility and reliability of the research.

Replay and regularization methods in CL prioritize computational efficiency, primarily because of their ability to leverage past experiences or impose constraints on model parameters without necessitating complex dynamic adjustments. Unlike dynamic methods, which often require additional model parameters or extensive memory usage to adapt to evolving tasks, replay methods can efficiently recycle past data samples for training. Similarly, regularization techniques impose constraints on the model weights based on previous tasks, effectively reducing the computational overhead associated with continual adaptation. This focus on computational efficiency ensures smoother integration into real-world applications, where resource constraints are prevalent, enabling CL models to adapt efficiently to new tasks over time.

\subsection{Practicality of CL}
\label{practicality}

In this section, we discuss the practical needs for CL in the medical domain and how the current research can be improved for real-life clinical deployment. We highlight four important practical needs of CL in the medical domain.
 
\subsubsection{Standardization}
The need for continual learning in the medical domain is paramount, mainly because of data scarcity and privacy concerns. However, current research in this domain lacks a clear structure and standard methodology for conducting experiments. This eventually makes it difficult to compare different studies, which may not provide a complete analysis of the proposed methods. Moreover, the absence of standardized benchmarks further complicates the validation of CL algorithms in medical imaging. Consequently, there is a pressing need to establish common evaluation protocols and datasets to ensure fair comparison and robust assessment of CL techniques in real-world healthcare settings. Furthermore, in the medical domain, the scarcity of publicly available datasets is a significant obstacle. The primary dataset used in recent studies is the MedMNIST dataset \citep{yang2021medmnist}. This dataset lacks a specific task structure, and different authors often choose different tasks and settings to evaluate their methods. Notably, even a small shift in the order of the tasks can cause a significant difference in overall performance. While most of the reviewed studies mention how they partitioned the datasets they used, the comparison becomes challenging owing to differences in settings, leading to unfair comparisons. Moreover, the size of the images in this dataset makes it debatable whether it's a benchmark that would transfer to clinical usability. Therefore, there is a clear need to develop standard datasets, comparison settings, and metrics to evaluate and compare the CL methods.

\subsubsection{Task Boundaries}
A crucial aspect to consider in CL is how task labels are assigned to different settings. For instance, if data are acquired from distinct scanners A, B, and C in a hospital where the source of each data point is known, a practical approach would involve training the network using the task label information corresponding to each scanner. In this scenario, clear task boundaries exist because we have information on the scanner used at each instance during the inference or deployment phase. However, a more practical clinical scenario involves a gradual shift in the data distribution. These shifts could arise from changes in image acquisition protocols and machines or from shifts in patient demographics, such as a gradual transition to an older patient population. These changes occur gradually and may not be immediately evident, and present a challenge with smooth task boundaries, where task labels are unknown during inference. In such scenarios, traditional CL methods that rely on fixed task labels fall short. Instead, adaptable algorithms are required to handle the evolving data distributions without predefined labels. Task-agnostic CL methods can alleviate this issue, and the goal of CL involves detecting the task label itself.

\subsubsection{Reproducibility and Computational Constraints}
Reproducibility and making the source code public are highly important in CL in the medical domain. This ensures that the findings and models could be validated and used by other researchers to foster trust and confidence in the results. The open-source development of these methods is crucial for the advancement of CL as a field. Moreover, the computational aspect is important for scalability and efficiency of systems for larger populations. Unlike the natural dataset domain, medical domain models must be deployed on-site in the hospital so that they can access the local clinical data repository without any privacy risk. The computational resources at hospitals are very limited compared with the large cloud-based solutions used in other domains, further increasing the need for efficient solutions. Therefore, reproducibility and computational efficiency are essential for advancing CL in the medical domain and ultimately improving patient care and outcomes.

\subsubsection{Regulatory Framework for Model Updates}

Several regulations govern update protocols for deployed models \citep{moskalenko2024resilience}. These regulations demand transparency in the update process, while ensuring that the model's performance does not deteriorate significantly on previously learned tasks. In addition to transparency and performance preservation, compliance with regulatory requirements often requires addressing various aspects. Ensuring data privacy is paramount, with updates required to strictly adhere to data privacy regulations to safeguard sensitive information. Ethical considerations, including fairness, bias, and accountability, must also be considered when updating the models to maintain ethical standards. Moreover, rigorous validation procedures are essential to assess the impact of updates on model performance and to ensure compliance with regulatory standards.

\section{Recommendation}
\label{recommendation}

Although significant strides have been made in mitigating catastrophic forgetting within the realm of CL in the medical domain, several critical research avenues remain unexplored, needing attention and investigation in the future. Based on our analysis of the existing literature, the following are some of the open challenges and possible future directions for CL in the medical field.

\textbf{Specialized CL Datasets in the Medical Domain}: There remains a pressing need for specialized datasets tailored to the unique demands of CL scenarios. These datasets would serve as invaluable resources, enabling researchers to advance methodologies within this specialized field. By providing access to data specifically designed for CL tasks, researchers can better understand and address challenges unique to ongoing learning in medical contexts.

\textbf{Accessible Code Repositories for CL Methodologies}: In the ongoing exploration of CL methodologies in the medical field, the availability of easily accessible and replicable code repositories emerges as a critical factor for advancing research endeavors. Accessible code repositories would not only streamline the research process but also facilitate the validation and comparison of various CL approaches. By providing researchers with readily available codebases, the scientific community can foster collaboration and accelerate progress in the development and refinement of CL techniques for medical applications.

\textbf{Emphasizing Explainability and Interpretability in CL Models}: As the development of CL models gains momentum in the medical domain, the emphasis on explainability and interpretability becomes increasingly crucial. In healthcare settings, where transparency and trust are paramount, the ability to understand and interpret model decisions is essential to gaining acceptance from stakeholders. By prioritizing the interpretability of CL models, researchers can enhance their utility and applicability in real-world healthcare contexts, thereby facilitating their integration into clinical practice and decision-making processes.

\textbf{Practical Deployment Strategies for CL Models in Healthcare}: In the endeavor to translate research advancements in CL into tangible benefits for healthcare, the development of practical deployment strategies emerges as a critical consideration. Efforts should be directed towards devising deployment protocols that align with regulatory frameworks and accommodate the timely and appropriate updating of CL models. By addressing deployment challenges such as regulatory compliance and scalability, researchers can bridge the gap between theoretical advancements and their real-world implementation, ultimately facilitating the adoption of CL models in healthcare settings.

\section{Conclusion}
\label{conclusion}

In practical medical scenarios, there is often a continuous stream of incoming data that requires models to be updated incrementally without encountering issues such as catastrophic forgetting. This paper presents an up-to-date and thorough overview of continual learning applications in the medical domain, encompassing the latest advancements in theory, methodology, and real-world implementation. Our approach emphasizes practicality and deployment considerations. It is promising to note the increasing and widespread interest in continual learning in the medical field, as observed across the broader AI community. This growing interest brings new insights and solutions to address unique medical challenges.

\bibliographystyle{IEEEtran}
\bibliography{refs}

\vfill

\end{document}